\documentclass[letter]{llncs}

\usepackage{amssymb}
\usepackage{graphicx}
\usepackage{enumerate}
\usepackage{hyperref}
\usepackage{url}
\usepackage{color,colortbl}
\definecolor{Gray}{gray}{0.9}

\usepackage{pdflscape}

\usepackage[ruled,vlined]{algorithm2e}

\usepackage{amssymb}
\usepackage{fca}
\usepackage{amsmath}
\usepackage{amsfonts}
\usepackage{alltt}

\usepackage{url}
\usepackage{amsmath}
\urldef{\mailsa}\path|dignatov@hse.ru|
\newcommand{\keywords}[1]{\par\addvspace\baselineskip
\noindent\keywordname\enspace\ignorespaces#1}

\usepackage{fca}

\begin{document}

\mainmatter  

\titlerunning{Towards a Unified Taxonomy of Biclustering Methods}
\authorrunning{Dmitry I. Ignatov \and Bruce W. Watson}

\title{Towards a Unified Taxonomy of Biclustering Methods}

%
%
\author{Dmitry I. Ignatov$^{1}$ \and Bruce Watson$^{2}$}
\authorrunning{}

\institute{$^{1}$National Research University Higher School of Economics, Moscow, Russia\
\mailsa\\
$^{2}$Stellenbosch University, South Africa
}

%
%

\toctitle{}
\tocauthor{}
\maketitle

\begin{abstract}
Being an unsupervised machine learning and data mining technique, biclustering and its multimodal extensions are becoming popular tools for analysing object-attribute data in different domains. Apart from conventional clustering techniques, biclustering is searching for homogeneous groups of objects while keeping their common description, e.g., in binary setting, their shared attributes. In bioinformatics, biclustering is used to find genes, which are active in a subset of situations, thus being candidates for biomarkers. However, the authors of those biclustering techniques that are popular in gene expression analysis, may overlook the existing methods. For instance, BiMax algorithm is aimed at finding biclusters, which are well-known for decades as formal concepts. Moreover, even if bioinformatics classify the biclustering methods according to reasonable domain-driven criteria, their classification taxonomies may be different from survey to survey and not full as well. So, in this paper we propose to use concept lattices as a tool for taxonomy building (in the biclustering domain) and attribute exploration as means for cross-domain taxonomy completion.

\end{abstract}

\section{Introduction}

Biclustering is a popular family of data analysis techniques within cluster-analysis. Previously biclustering was known under the names direct clustering or subspace clustering \cite{Hartigan:1972}. The term biclustering was proposed by Boris Mirkin in \cite{Mirkin:1996}, p. 296:

\noindent \textit{The term biclustering refers to simultaneous clustering
of both row and column sets in a data matrix. Biclustering addresses the problems of aggregate representation of the basic features of interrelation
between rows and columns as expressed in the data.}

The main advantage of biclustering technique lies in its ability to keep similarity of grouped objects in terms of their common attributes. So, biclustering is able to capture object similarity (homogeneity) expressed only by a subset of attributes, which allows an analyst to clearly see why certain objects were grouped together.

In the previous decade biclustering methods became extremely popular for gene expression analysis analysis in bioinformatics. Here, genes which demonstrate similar properties only in a subset of observable situations are considered to be within a bicluster along with those situations. The first rather comprehensive survey in the field was done in \cite{Madeira:2004}. Even though the survey was limited only to biclustering in bioinformatics, the field came to its maturity to have its own classification of the methods. The authors classified biclustering techniques according to several properties: biclustering type, biclustering structure, the way of bicluster generation, and the algorithmic strategy.

As it often happens, researchers from the bioinformatics domain overlooked or even rediscovered biclustering methods which have been known for decades. Thus, the notion of formal concept was known since the early 80-s \cite{wille:1982}, it corresponds to maximal inclusion unit submatrices in Boolean matrices \cite{Ganter:1999,Barkow:2006,Kaytoue:2009}. The idea of closed sets from Formal Concept Analysis and from frequent itemset mining were not considered in the bioinformatics domain. However, there are numerous efficient algorithms and applications, which can be treated as special cases of biclustering-based ones. To the best of our knowledge there is no any biclustering technique mentioned in bioinformatics which exploits ordered bicluster hierarchies.  Thus, in Formal Concept Analsis, biclusters (formal concepts) are hierarchically ordered by the relation ``be more general concept than'', which proved its helpfulness for data exploration and taxonomy building in different domains.

The aim of this work is two-fold: on the one hand, we are going to shed light on neighbouring domains where biclustering is actively used, and on the other hand build lattice-based taxonomies using the existing classifications of biclustering algorithms in the literature. The main open question in this work is as follows: How to build a unified taxonomy of the biclustering techniques.

The rest of the paper is organised as follows. In Section \ref{sec:prev}, we shortly review previous work on biclustering, taxonomies of algorithms, and related fields. In Section \ref{sec:basics} we give basic definitions of FCA and biclustering (in the most general form). In Section \ref{sec:multi} we outline several existing biclustering extensions under the the name of multimodal clustering.
 Section \ref{sec:tax} is the main part of the paper which provides examples of different taxonomies of biclustering algorithms obtained from the literature.

\section{Previous work}\label{sec:prev}

Construction of taxonomies of algorithms in Computer Science is not new. Thus, in \cite{Cleophas:2005,Cleophas:2006} a taxonomy of string matching algorithms was built guided by domain experts according to TABASCO methodology. In those papers, it was shown that concept lattices can be a good visualisation tool paired with interactive abilities of modern computer software. Moreover, concept lattices were successfully used for epistemic taxonomy building \cite{Roth:2008} combining multiple inheritance feature with compact graphical representation.

In addition to the existing term biclustering, there are several others like co-clustering or simultaneous clustering. 
Triclustering, Triadic FCA, multimodal clustering, clustering of Boolean tensors, closed n-sets, relational clustering and several other techniqes are all examples of possible biclustering extensions.  

\section{Basic definitions and models}\label{sec:basics}

\subsection{Formal Concept Analysis}\label{ssec:FCA}

\begin{definition} A \textbf{formal context} $\context=(G,M,I)$ consists of two sets $G$ and $M$ and a relation  $I$ between $G$ and $M$. The elements of $G$ are called the \textbf{objects} and the elements of  $M$ are called the \textbf{attributes }of the context. The notation $gIm$ or $(g,m) \in I$ means that the object $g$ has attribute $m$. 
\end{definition} 

\begin{definition} 
	
	For $A \subseteq G$, let
	
	$$A^\prime := \{m \in M | (g,m) \in I \mbox{ for all } g \in A\}$$
	
	and, for $B \subseteq M$, let  
	
	$$B^\prime := \{ g \in G | (g,m) \in I \mbox{ for all } m \in B\}.$$
	
	These operators are called \textbf{derivation operators} or \textbf{ concept-forming operators} for $\context=(G,M,I)$. 
	
\end{definition}

\begin{proposition} Let $(G,M,I)$ be a formal context, for subsets $A, A_1, A_2 \subseteq G$ and $B\subseteq M$ we have
	\label{primeprop}
	\begin{enumerate}
		\item $A_1 \subseteq A_2$ iff $A_2^\prime \subseteq A_1^\prime$,
		\item $A \subseteq A^{\prime\prime}$,
		\item $A = A^{\prime\prime\prime}$ (hence, $A'''' = A''$),
		\item $(A_1 \cup A_2)' = A'_1 \cap A'_2$,
		\item $A\subseteq  B'\Leftrightarrow B\subseteq A' \Leftrightarrow A\times B \subseteq I$.
	\end{enumerate}
	
	Similar properties hold for subsets of attributes.
	
\end{proposition}

\begin{definition}
	
	A \textbf{closure operator} on set $S$ is a mapping $\varphi \colon 2^S \to 2^S$ with the following properties:
	
	\begin{enumerate}
		\item[1.] ${\varphi{\varphi X}} = {\varphi X}$ (\textbf{idempotency})
		
		\item[2.] $X\subseteq {\varphi X}$ (\textbf{extensity})
		
		\item[3.] $X\subseteq Y\Rightarrow {\varphi X}\subseteq {\varphi Y}$ (\textbf{monotonicity})
	\end{enumerate}
	
	For a closure operator $\varphi$ the set  $\varphi X$ is called \textbf{closure} of $X$.
	
	A subset $X\subseteq G$ is called \textbf{closed} if ${\varphi X} = X$.
	
\end{definition}

Let $(G,M,I)$ be a context, one can prove that operators
	$$(\cdot)''\colon 2^G\to 2^G,\ (\cdot)''\colon 2^M\to 2^M$$
	are closure operators. 

\begin{definition} 
	A \textbf{formal concept} of a formal context $\context=(G,M,I)$ is a pair $(A,B)$ with $A \subseteq G$,
	$B \subseteq M$, $A^\prime = B$ and $B^\prime = A$. The sets $A$ and $B$ are called the extent and the intent
	of the formal concept $(A,B)$, respectively. The \textbf{subconcept-superconcept relation} is
	given by $(A_1,B_1) \leq (A_2,B_2)$ iff $A_1 \subseteq A_2$ ($B_1 \subseteq B_2$).
\end{definition} 

This definition says that every formal concept has two parts, namely, its extent and
intent. This follows an old tradition of the \textit{Logic of Port Royal (1662)}, and is in line with the International Standard ISO 704 that formulates the following definition: ``A concept is considered to be a unit of thought constituted of two parts: its extent and its intent.''

\begin{definition} The set of all formal concepts of a context $\context$ together with the order relation $I$ forms a complete lattice, called the \textbf{concept lattice } of $\context$ and denoted by $\BV(\context)$.
\end{definition}

\begin{definition}
\textbf{Implication} $A\to B$, where $A, B\subseteq M$ holds in context $(G,M,I)$ if
$A'\subseteq B'$, i.e., each object having all attributes from
$A$ also has all attributes from  $B$.
\end{definition}

\subsection{Biclustering}\label{ssec:bic}

In the first survey on biclustering techniques \cite{Madeira:2004}, bicluster is defined as a submatrix of an input object-attribute matrix. That is for a given matrix $A \in \mathbb{R}^{m\times n}$, a bicluster $b$ is a pair $(X,Y)$, where $X \subseteq \{1,\cdots,m \}$ and $Y \subseteq \{1,\cdots, m \}$. The bicluster should fulfil a certain homogeneity property, which varies from method to method, e.g., it may be allowed to contain only 1s inside the corresponding submatrix (bicluster) in Boolean case.

For instance, for analysing large markets of context advertisement, we propose the following FCA-based definition of a bicluster \cite{Ignatov:2010,Ignatov:2012b}.

\begin{definition}\label{def:bicl}
	If $(g,m)\in I$, then $(m',g')$ is called an object-attribute or \emph{OA-bicluster}
	with density $\rho(m',g')=\frac{|I\cap (m'\times g')|}{|m'|\cdot|g'|}$.
	
\end{definition}

Here are some basic properties of oa-biclusters.

\begin{proposition}\label{proposition:bicl}
	
\
	
	1. $0\leq \rho\leq 1$.
	
	2. oa-bicluster $(m',g')$ is a formal concept iff $\rho = 1$.
	
	3. if $(m',g')$ is a oa-bicluster, then $(g'',g')\leq (m',m'')$.
	
\end{proposition}

\begin{figure}[h]
	\begin{center}
		\includegraphics[scale=0.5]{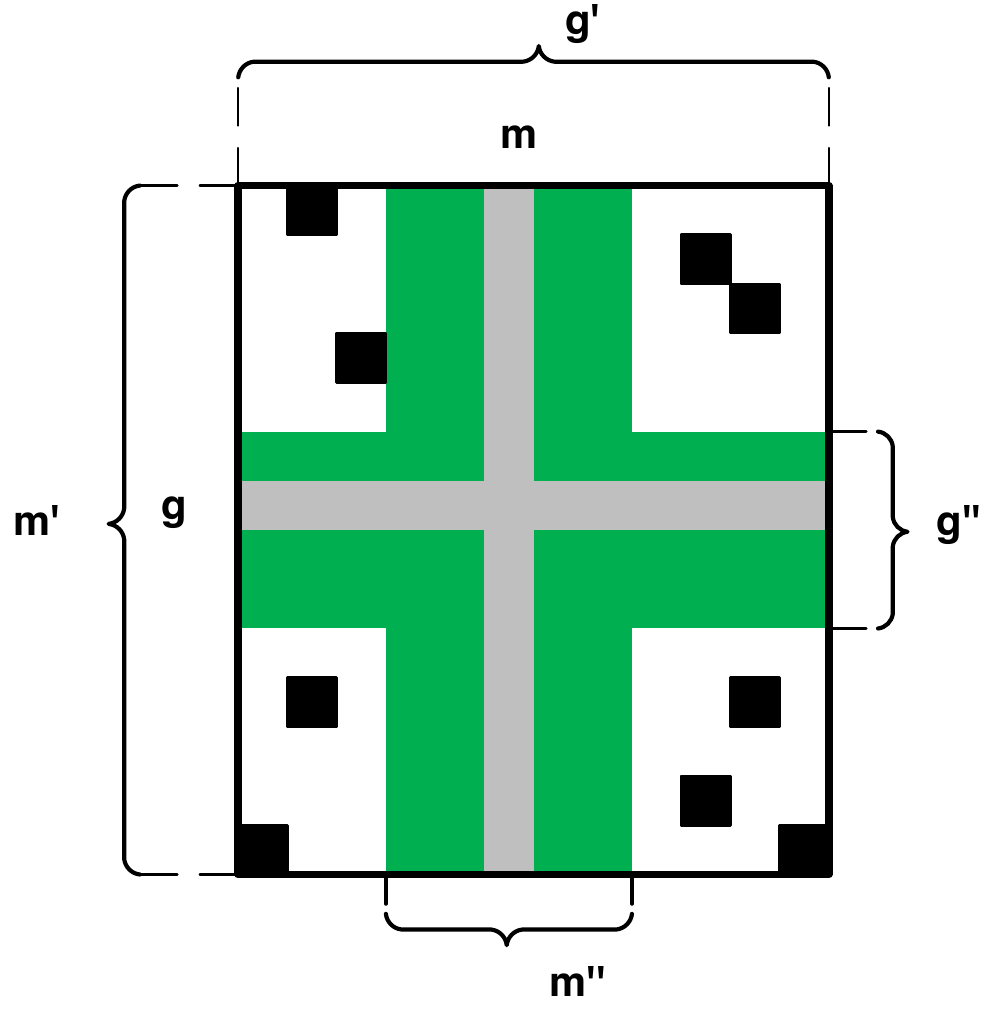}
		\vskip -0.1in
		\caption{OA-bicluster based on object and attribute closures}\label{fig:Bicl}
	\end{center}
	\vskip -0.1in
\end{figure}

In figure \ref{fig:Bicl} you can see the example of the oa-bicluster for a particular pair $(g,m) \in I$ of a certain context $(G,M,I)$. In general, only the regions $(g'',g')$ and $(m',m'')$ are full of non-empty pairs, i.e. have maximal density $\rho=1$, since they are object and attribute formal concepts respectively. Some black cells indicate non-empty pairs which one may found in such a bicluster. Therefore, the density parameter $\rho$ is a bicluster quality measure which shows how many non-empty pairs the bicluster contains.

\begin{definition}\label{def:dense}
	Let  $(A,B) \in 2^G \times 2^M$ be a oa-bicluster and  $\rho_{min}$ be a nonnegative real number, such that $0 \leq \rho_{min} \leq 1$, then $(A,B)$ is called  \emph{dense} if it satisfies the constraint $\rho(A,B)\geq \rho_{min}$.
\end{definition}

\section{Existing taxonomies and their analysis}\label{sec:tax}

\begin{figure}
	\centering
	\includegraphics[width=1.00\textwidth]{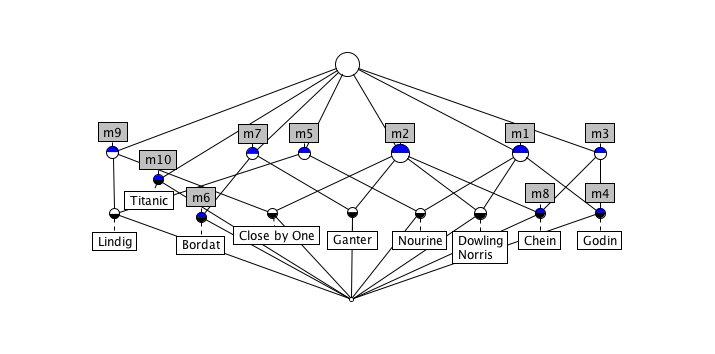}
	\caption{The line diagram of concept lattice for FCA algorithms}
	\label{fig:tax2002}
\end{figure}

Since formal concept is a natural notion of bicluster for Boolean data and was rediscovered or reused in bioinformatics, one may suppose that the taxonomy of FCA algorithms is a part of the taxonomy of biclustering algorithms. In fact, paper \cite{Kuznetsov:2002} proposed such a taxonomy (see fig.\ref{fig:tax2002}).

The classification properties of the concept lattice building algorithms encoded as follows: 

\begin{itemize}

\item  $m1$ means incremental approach;
\item $m2$ means that an algorithm uses canonicity based on the lexical order;
\item $m3$ means that an algorithm divides the set of concepts into several parts; 
\item $m4$ designates that an algorithm uses hashing;
\item $m5$ means that an algorithm maintains an auxiliary tree structure; 
\item $m6$ means usage of attribute cache; 
\item $m7$ encodes that an algorithm computes intents by subsequently computing intersections of object intents (i.e., $\{g\}' \cap \{h\}'$);
\item  $m8$ means that an algorithm computes intersections of already generated intents; 
\item $m9$ encodes that an algorithm computes intersections of non-object intents and object intents;
\item $m10$ means that an algorithm uses supports of attribute sets.

\end{itemize}

\begin{landscape}
	
	\begin{figure}
		\centering
		\includegraphics[width=1.8\textwidth]{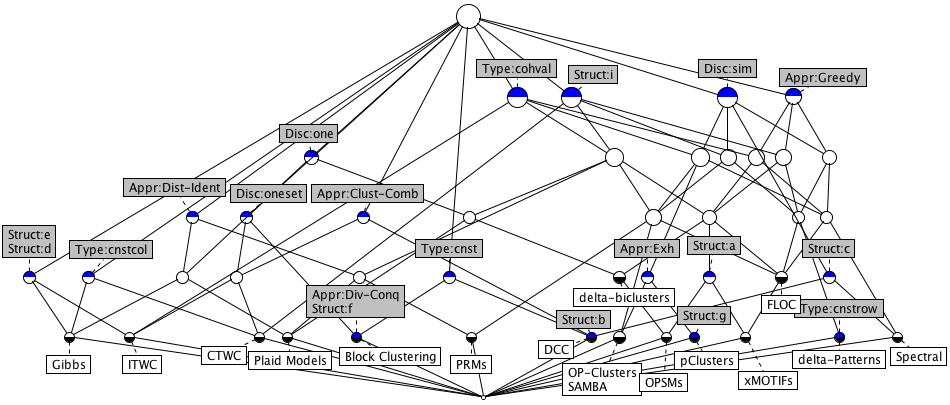}
		\caption{The line diagram of concept lattice for the biclustering taxonomy of Madeira and Oliveira (2004)}
		\label{fig:tax2004}
	\end{figure}
	
\end{landscape}

We formed a context based on Table II
``Overall comparison of the biclustering algorithms'' \cite{Madeira:2004} and build its concept-based taxonomy in Fig.˜\ref{fig:tax2004}.

Originally, the authors used several criteria to classify the existing (reviewed) biclustering algorithm: type of bicluster, structure of biclusters, type of bicluster discovery, and algorithmic strategy. 

Thus, with respect to the definition of bicluster (its type) the authors differentiate between 1) biclusters with constant values, 2) biclusters with constant values on rows or columns, 3) biclusters with coherent values, and 4) biclusters with coherent evolutions.

The biclusters were classified into one of 9 classes according to their structure.

\begin{itemize}
	\item[a)] Single Bicluster
	\item[b)] Exclusive row and column biclusters (rectangular diagonal blocks after row and column reorder).
	\item[c)] Non-Overlapping biclusters with checkerboard structure.
	\item[d)] Exclusive-rows biclusters.
	\item[e)] Exclusive-columns biclusters.
	\item[f)] Non-Overlapping biclusters with tree structure.
	\item[g)] Non-Overlapping non-exclusive biclusters.
	\item[h)] Overlapping biclusters with hierarchical structure. 
	\item[i)] Arbitrarily positioned overlapping biclusters.
\end{itemize}

Different biclustering methods pursue different goals in terms of the number of discovered biclusters. Thus, they may identify one bicluster at a time or be targeted to discovering one set of biclusters at a time, or they can follow simultaneous bicluster identification, which means that the biclusters are discovered all at the same time. All the three types are possible values of Discovery type attribute in the proposed taxonomy.

Since in many cases the biclustering enumeration is a hard task (the corresponding counting problem may belong to $\#P$ complexity class), different algorithmic enumeration strategies were proposed. Thus, Madeira and Oliviera sort out several categories: 1) Iterative Row and Column Clustering Combination, 2) Divide and Conquer, 3) Greedy Iterative Search, 4) Exhaustive Bicluster Enumeration, and 5) Distribution Parameter Identification.

One of the reasonable questions here is: Why should we build diagrams instead of looking at tables? The answer is we need two complementary views, object-attribute descriptions in tables and ordered clusters of objects such that the objects inside a particular cluster (formal concept) share the same attributes. It is not easy to find such clusters with respect to permutations of rows and columns manually even for small contexts. Moreover, by examining the concept lattice of a certain taxonomy we can find useful attribute dependencies, which can help to discover the underlying taxonomy's domain.

\begin{landscape}
	
	\begin{figure}
		\centering
		\includegraphics[width=1.8\textwidth]{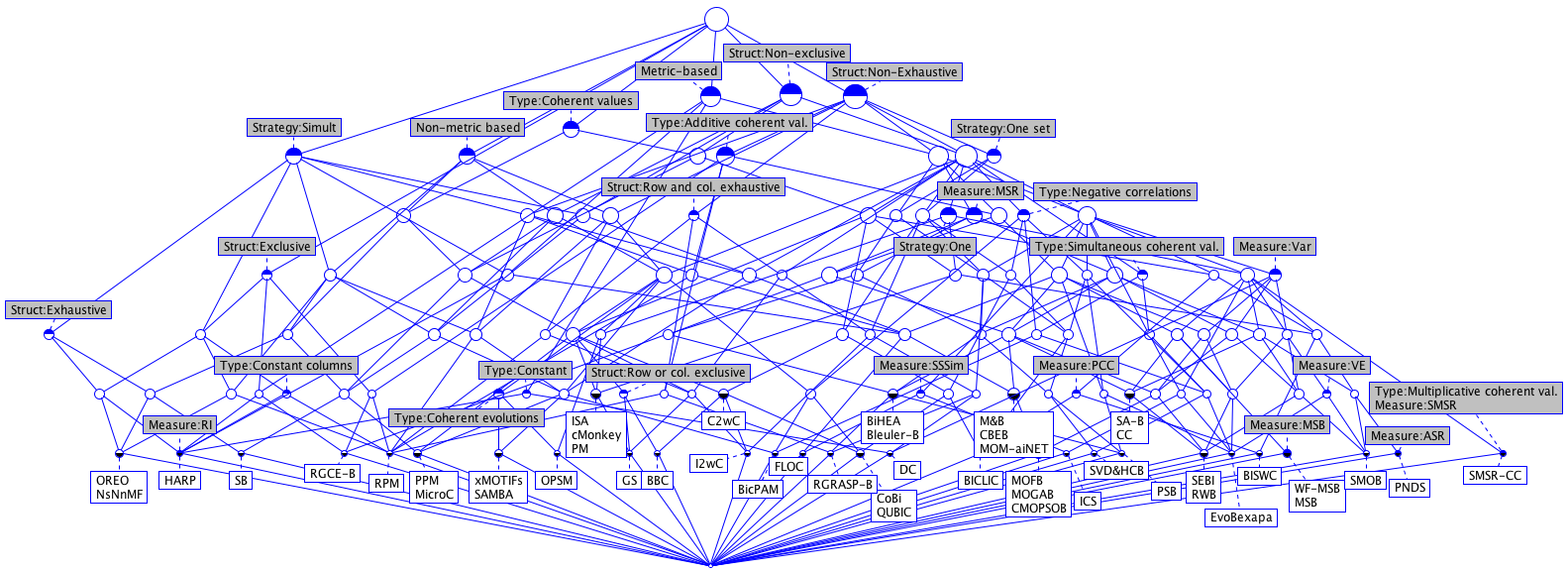}
		\caption{The line diagram of concept lattice for the biclustering taxonomy of Pontes et al. (2015)}
		\label{fig:tax2015}
	\end{figure}
	
\end{landscape}

The previous classification done by Madeira and Oliviera was extended and completed almost 11 years later in \cite{Pontes:2015}. We build the corresponding line diagram in Fig.˜\ref{fig:tax2015}.

In fact, the number of classified methods were extended to 47 from 16.

The authors slightly redesigned the proposed classification criteria.  Thus, they split the analysed methods into two categories: metric-based and non-metric based. We counted this split as two corresponding attributes in the related formal context. However, we also decided to include all the mentioned evaluation metrics into our analysis like ``Measure:MSR'' meaning Mean Squared Residue.

The remaining criteria have been changed or extended by the authors. For instance, instead of bicluster types, now eight patterns has been proposed:
\begin{enumerate}
\item Constant;
\item Constant columns;
\item Coherent values;
\item Additive coherent values;
\item Multiplicative coherent values;
\item Simultaneous coherent values;
\item Coherent evolutions;
\item Negative correlations.
\end{enumerate}

The sub-taxonomy based on bicluster structure now contains only six criteria: row exhaustive, column exhaustive, non-exhaustive, row exclusive, column exclusive, and non-exclusive.
 
By means of terms ``exhaustive'' and ``exclusive'' it is possible to describe the desired structure. Thus, exhaustive means where all genes (conditions) should belong to some bicluster, i.e. to be covered by it. Exclusive means whether a gene (condition) has to belong no more than one bicluster; e.g., in non-exclusive case overlapping is allowed.

The attribute algorithmic strategy has been altered in its original form from \cite{Madeira:2004}. The attribute Discovery from \cite{Madeira:2004} has been renamed to Strategy, but the values remain the same: one bicluster at a time, set of biclusters at a time, and simultaneous bicluster identification.

In the beginning of 2000s it was unusual that data analysts and biologists can miss existing biclustering methods (like FCA), which were not applied in the bioinformatics domain yet. However, later FCA was successfully applied in the domain of gene expression analysis \cite{Besson:2005,Kaytoue:2008,Motameny:2008,Kaytoue:2009,Kaytoue:2011}, formal concepts were rediscovered by \cite{Barkow:2006,Prelic:2006} in bioinformatics, approximate greedy \cite{Mirkin:2011} and fast \cite{Ignatov:2008,Ignatov:2012b} methods for dense bicluster discovery in Boolean setting appeared.

However, even the recent taxonomy from \cite{Pontes:2015} does not include any of them. 

To overcome incompleteness caused by the bioinformatics domain view restriction, an attempt to extend the taxonomy of Madeira and Oliveira was done in \cite{Ignatov:2010}.

	\begin{figure}
		\centering
		\includegraphics[width=1\textwidth]{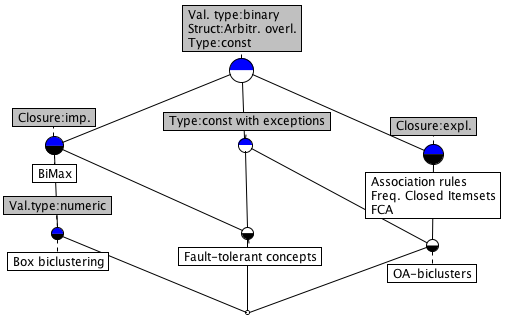}
		\caption{The line diagram of concept lattice for the biclustering taxonomy of Ignatov (2010)}
		\label{fig:tax2010}
	\end{figure}

In addition to the existing criteria, the attribute bicluster values type was added taking two values: binary and numeric. The former shows whether the method is able to find patterns in Boolean object-attribute tables and the latter indicates whether the input entries from $\mathbb{R}$ can be processed by an algorithm. Another important criteria is whether an algorithm based on the notion of closure (operator) from FCA and Closed Frequent Itemset Mining, implicitly or explicitly. The corresponding formal context is given below.

\begin{center}
	\begin{cxt}%
		\cxtName{FCA-related biclustering}%
		\atr{Type:const}
		\atr{Type:const with exceptions}
		\atr{Struct:Arbitr. overl.}
		\atr{Value type:binary}
		\atr{Closure:explicit}
		\atr{Closure:implicit}
		\atr{Val.type:numeric}
		\obj{X.XX.X.}{BiMax}
		\obj{X.XX.XX}{Box biclustering}
		\obj{X.XXX..}{FCA}
		\obj{X.XXX..}{Freq. Closed Itemsets}
		\obj{X.XXX..}{Association rules}
		\obj{XXXX.X.}{Fault-tolerant concepts}
		\obj{XXXXX..}{OA-biclusters}			
	\end{cxt}
\end{center}

In Fig.\ref{fig:tax2010} we show only those objects (biclustering algorithms)  which were not included in the taxonomy of Madeira and Oliveira. By so doing, we considered three general approaches (Formal Concept Analysis, Frequent Closed Itemsets, and Association Rules) and particular modifications of formal concepts (box biclusters \cite{Mirkin:2011}, fault-tolerant concepts \cite{Pensa:2005}, OA-biclusters \cite{Ignatov:2010,Ignatov:2012b}).

In \cite{Ignatov:2010} it was noted that every association rule of the form $A \rightarrow B$ corresponds to the bicluster $(A' \cup B', A \cup B)$ or $(A' \cap B', A \cup B)$, hence it is possible to consider association rules as a biclustering technique.

Later, the taxonomical issues of biclustering were discussed in \cite{Flores:2013}; the authors sorted out binary, integer, and real-valued biclustering approaches. Frequent itemset mining algorithms, BiMax, Association Rules and several others were included into a recent survey on pattern-based biclustering algorithms  \cite{Henriques:2015}. Even though the idea of closure was mentioned by the survey authors, they did not use closedness property of the studied patterns as a taxonomical attribute.	Moreover, even though the authors cited several FCA-based papers ob biclustering, they do not treat relationship of this discipline with biclustering. However, as we have mentioned, such relationships were studied on a solid mathematical and experimental level; for example, in \cite{Kaytoue:2014} it was shown that biclustering of numerical data is possible by means of triclustering of the corresponding binary relation.

Implications and Attribute Exploration can help to find hidden attribute dependencies and extend the built taxonomy by new (counter)examples.

Below we provide the reader with top-10 (w.r.t. support measure) implications of Duquenne-Gigues base \cite{Guigues:1986} derived from the formal context of the taxonomy from \cite{Pontes:2015}.

\begin{itemize}
\item   $\{ \mbox{Metric-based, Struct:Non-exclusive} \} \rightarrow \{ \mbox{Struct:Non-Exhaustive} \}$, $sup= 26$
\item  $\{\mbox{Type:Additive coherent val.} \} \rightarrow \{ \mbox{Struct:Non-Exhaustive}\}$,  $sup= 20$
\item $\{\mbox{Measure:MSR} \} \rightarrow \{ \mbox{Metric-based, Struct:Non-Exhaustive}\}$, $sup= 18$ 
\item  $\{\mbox{Type:Additive coherent val., Struct:Non-Exhaustive, Struct:Non-exclusive} \} \rightarrow \{ \mbox{Metric-based}\}$, $sup= 18$
\item  $\{\mbox{Strategy:One} \} \rightarrow \{ \mbox{Struct:Non-Exhaustive, Struct:Non-exclusive}\}$, $sup= 17$
\item $\{\mbox{Type:Coherent values, Struct:Non-Exhaustive} \} \rightarrow \{ \mbox{Struct:Non-exclusive}\}$, $sup= 15$
\item  $\{\mbox{Strategy:One set} \} \rightarrow \{ \mbox{Struct:Non-Exhaustive}\}$, $sup= 13$ 
\item  $\{\mbox{Measure:Var} \} \rightarrow \{ \mbox{Metric-based, Struct:Non-Exhaustive, Struct:Non-exclusive}\}$, $sup= 8$
\item $\{\mbox{Type:Negative correlations} \} \rightarrow \{ \mbox{Struct:Non-Exhaustive, Struct:Non-exclusive}\}$,  $sup= 7$
\item  $\{\mbox{Metric-based, Struct:Non-Exhaustive, Strategy:Simult} \} \rightarrow$

 $ \{ \mbox{Type:Additive coherent val., Struct:Non-exclusive}\}$, $sup= 7$ 
\end{itemize}

Since we deal with implications, their confidence measure is equal to 1. The size of the whole set of implications in Duquenne-Gigues base is 105.

If we start attribute exploration for the same context, then the first question in a row is the following:

Is it true, that when biclustering technique has attribute ``Strategy:One set'', that it also has attribute ``Struct:Non-exhaustive''?

An expert can either agree with the implication  $\{ \mbox{Strategy:One set} \to \mbox{Struct:Non-exhaustive}\}$ or disagree. In the latter case, (s)he needs to provide a counterexample: a biclustering technique which follows discovery strategy ``one set of biclusters at a time'' but does not result in biclusters of the structure type ``exhaustive''. There is also an option to stop Attribute Exploration process at every step.

\section{Multimodal clustering and closed n-sets}\label{sec:multi}

Since the field of biclustering is a subdomain of multimodal or relational clustering, the taxonomy can be extended by applying similar criteria to n-ary relation and tensor clustering algorithms.

Thus, the notion of formal concept was generalised for triadic \cite{Lehmann:1995} and polyadic case \cite{Voutsadakis:2002}. There are efficient algorithms to find triconcepts \cite{Jaschke:2006} and poliadic concepts (closed n-sets) \cite{Cerf:2009}. There exist relaxations of triconcept and poliadic concept notions, triclusters and n-clusters, which allow for certain entries inside such n-dic concept to be zeros \cite{Mirkin:2011,Ignatov:2013,Cerf:2013}; the theoretical and experimental comparison is done in \cite{Ignatov:2015ml}.  There are also methods for mining closed patterns in n-ary relations \cite{Spyropoulou:2014}. Two biclustering approaches can be used for mining two formal contexts simultaneously, which shares either set of attributes or objects; this results in pseudotriclusters \cite{Gnatyshak:2012}. As for purely biological applications of triclustering we may suggest reading, for example, \cite{Zhao:2005} and \cite{Li:2009}.

\section{Conclusion and future work}

Even though the taxonomy building of a particular subfield of Data Analysis or Computer Science is not as laborious as devising Carl Linnaeus' pre-phylogenetic taxonomy, this is not an easy task to merge several such existing taxonomies and build a unified one. Similarly to new species discovery, new algorithms can be proposed and since they can contain new specific features, new classification attributes may be needed.

One of the possible schemes of taxonomy maintaining here could be done in terms of Attribute Exploration \cite{Ganter:1999}.

At a certain moment the group of expert fixes the set of existing biclustering methods and proposes suitable criteria for their classification. A person or a team which has proposed a new biclustering method should classify the method according to the chosen scheme, then it should be validated by experts. Such a team can propose an extra criterion for method classification. If a person or a team is going to propose a new method for an unexplored combination of classification attributes, it is possible to run attribute exploration to see which prospective types of methods are missing to date.  By means of Object Exploration, it may become clear that some attributes are missing, e.g. it is evident that formal concepts or Boolean biclusters is only a particular case of bicluster type with constant values and we need at least one new attribute, Boolean entry values.

Since a taxonomy may be used not only for classification itself, but as a search index for potential users, we may suggest using several ways of interactive visualisation: tree-based (TABASCO-like), concept lattice based (line diagrams), object-attribute tables, and nested line diagrams. The latter can help when someone is interested in a special main set of attributes, which should be shown in the outer taxonomy on the line diagram; the inner taxonomy can be shown if the method-seeker needs a finer level granularity or more detailed description inside of the selected node from the outer taxonomy. 

It is important to note that taxonomies can be considered as a special case of ontologies, and here FCA was successfully used both for ontology merging and completion \cite{Ganter:03,Sertkaya:2010}.

There are two main tasks for our future studies: 1) unifying the existing bicluster taxonomies, and 2) creation a taxonomy of multimodal clustering techniques. Even though there are several good tools for building and managing concept lattices like Concept Explorer, we need to rely on more flexible tools with extensible components. In particular we hope that FCART can become our tool of choice in the near future \cite{Neznanov:2014}.

\subsubsection*{Acknowledgements.}
We would like to thank  Sergei Obiedkov and Derrick Kourie for a piece of advice and their earlier work on the topic. This work was supported by the Basic Research Program at the National Research University Higher School of Economics in 2015-2016 and performed in the Laboratory of Intelligent Systems and Structural Analysis. The first author was also supported by Russian Foundation for Basic Research (grant \#13-07-00504).

\bibliographystyle{splncs}

\bibliography{bictax}
\end{document}